\documentclass{article}
\usepackage[usenames,dvipsnames]{xcolor}
\usepackage{jfrExamplee}
\usepackage{graphicx}
\usepackage{svg}
\usepackage{textcomp}
\graphicspath{{./figures/}}
\usepackage{apalike}
\usepackage{setspace}
\usepackage{amsmath}
\usepackage{siunitx}
\usepackage{enumitem}[itemsep=1p]
\usepackage{url}
\usepackage{algorithm}
\usepackage{algpseudocode}   

\title{AgriCruiser: An Open Source Agriculture Robot for Over-the-row Navigation}

\author{
\hspace{-2em}Kenny Truong\\
\hspace{-2em}Dept. of Mechanical \& Aerospace Engineering\\
\hspace{-2em}University of California, Los Angeles\\
\hspace{-2em}\texttt{kennytruoong@g.ucla.edu} \\
\And
\hspace{-2em}Yongkyu Lee\thanks{Equal contribution} \\
\hspace{-2em}Dept. of Mechanical \& Aerospace Engineering\\
\hspace{-2em}University of California, Los Angeles\\
\hspace{-2em}\texttt{yongkyulee@g.ucla.edu} \\
\AND
Jason Irie\footnotemark[1] \\
Dept. of Mathematics\\
University of California, Los Angeles\\
\texttt{jasonirie@g.ucla.edu} \\
\And
Shivam Kumar Panda\footnotemark[1] \\
Dept. of Mechanical \& Aerospace Engineering\\
University of California, Los Angeles\\
\texttt{shivamkp@g.ucla.edu} \\
\AND
\hspace{-2em}Mohammad Jony\\
\hspace{-2em}Dept. of Plant Sciences\\
\hspace{-2em}North Dakota State University\\
\hspace{-2em}\texttt{mohammad.jony@ndsu.edu}\\
\And
Shahab Ahmad \\
Dept. of Plant Sciences \\ 
North Dakota State University \\
\texttt{shahab.ahmad@ndsu.edu} \\
\AND
Md. Mukhlesur Rahman\\
Dept. of Plant Sciences \\ 
North Dakota State University \\
\texttt{md.m.rahman@ndsu.edu} \\
\And
M. Khalid Jawed \\
Dept. of Mechanical \& Aerospace Engineering\\
University of California, Los Angeles\\
\texttt{khalidjm@seas.ucla.edu} \\
}

\begin{document}

\maketitle

\begin{abstract}

We present the AgriCruiser, an open-source over-the-row agricultural robot developed for low-cost deployment and rapid adaptation across diverse crops and row layouts. The chassis provides an adjustable track width of 1.42 m to 1.57 m, along with a ground clearance of 0.94 m. The AgriCruiser achieves compact pivot turns with radii of 0.71 m to 0.79 m, enabling efficient headland maneuvers. The platform is designed for the integration of the other subsystems, and in this study, a precision spraying system was implemented to assess its effectiveness in weed management. In twelve flax plots, a single robotic spray pass reduced total weed populations (pigweed and Venice mallow) by 24- to 42-fold compared to manual weeding in four flax plots, while also causing less crop damage. 
Mobility experiments conducted on concrete, asphalt, gravel, grass, and both wet and dry soil confirmed reliable traversal consistent with torque sizing. The complete chassis can be constructed from commodity T-slot extrusion with minimal machining, resulting in a bill of materials costing approximately \$5,000–\$6,000, which enables replication and customization. The mentioned results demonstrate that low-cost, reconfigurable over-the-row robots can achieve effective weed management with reduced crop damage and labor requirements, while providing a versatile foundation for phenotyping, sensing, and other agriculture applications. Design files and implementation details are released to accelerate research and adoption of modular agricultural robotics.
\end{abstract}

\section{Introduction}
In recent years, the agricultural industry has faced significant challenges, including labor shortages, rising labor costs, and a growing emphasis on environmental sustainability and worker safety \cite{labor_shortage}. 
A decline in the workforce has led to higher labor costs; according to the USDA, the inflation-adjusted hourly wage for farm workers increased by 28\% between 2000 and 2022 \cite{wage_increase}. As a result, farm owners who depend on manual labor face higher operational expenses and reduced profitability due to labor shortages. 
 
In addition, manual weed control has long been a labor-intensive and hazardous process that relies on the handling of pesticides, often exposing workers to toxic chemicals. Long-term direct contact with these substances has been linked to various health risks, including skin disorders and chronic diseases \cite{pesticides}. As agricultural tasks continue to depend heavily on manual labor, the safety of workers remains a growing concern. Therefore, precision spraying, that can minimize crop contact and herbicide drift, has been the subject of extensive research in the agricultural market \cite{weed_management_2}. 

Thus, agricultural robotics is being studied to tackle the aforementioned problems, including safety concerns, labor shortages, and wage increases. The incorporation of robotics has the potential to improve productivity, maximize resource utilization, and support sustainable farming methods \cite{advantage}. 

Existing ground robots and tractor‑mounted implements achieve high throughput but typically require fixed row spacings, ample headland room, or crop‑specific geometries. These robots are often designed for specific crop types or to perform single tasks (e.g., single‑purpose sprayers/weeders), making them less adaptable to diverse farming needs \cite{task-specific}. Most of the platforms lack the over‑the‑row clearance required to traverse multi‑crop plots through mid‑season growth without damaging the canopy. Although robots such as Warthog UGV from Clearpath Robotics \cite{warthog} or Thorvald from Saga Robotics \cite{thorvald} also offer dynamic structural reconfiguration; however, these commercial multipurpose agricultural robots often come with a high price tags. The cost can typically range from \$60,000 to \$100,000 or more, as demonstrated by the implementation of Thorvald platforms at the West Central Research and Outreach Center \cite{weed_terminator}. Their high cost poses a barrier to the wide-scale adoption of robotics in agricultural settings, particularly for small- to medium-sized farms and for research purposes. Low‑cost open‑hardware efforts, on the other hand, tend to sacrifice either clearance, efficiency, maneuverability, or documentation, limiting reproducibility and cross‑lab comparison \cite{vasconcelos2023low,farmbot-genesis,mini_robot}. Consequently, there is a gap for a reconfigurable, open, and empirically validated platform that can serve both as a field research tool and as a practical weeding aid.

In contrast, our platform is designed to offer both functional adaptability and cost efficiency. The chassis is reconfigurable to different terrain conditions, row widths, canopy heights, and crop types while keeping total build cost to \$5,000–\$6,000 and minimizing fabrication labor. Unlike traditional heavy machinery such as tractors, which can cause soil compaction and an elevated risk of crop damage \cite{tractors}, the AgriCruiser utilizes a large yet agile platform. Its compact design features in-place rotation, which improves navigation through narrow crop rows, minimizes crop damage, and allows for more targeted operations. Furthermore, by offering open source access to the platform, we promote innovation through community collaboration and allow other researchers, developers, and farmers to adapt and build on the robot for a variety of use cases. 

\begin{figure}[h!]
    \centering
    \includegraphics[width=\linewidth]{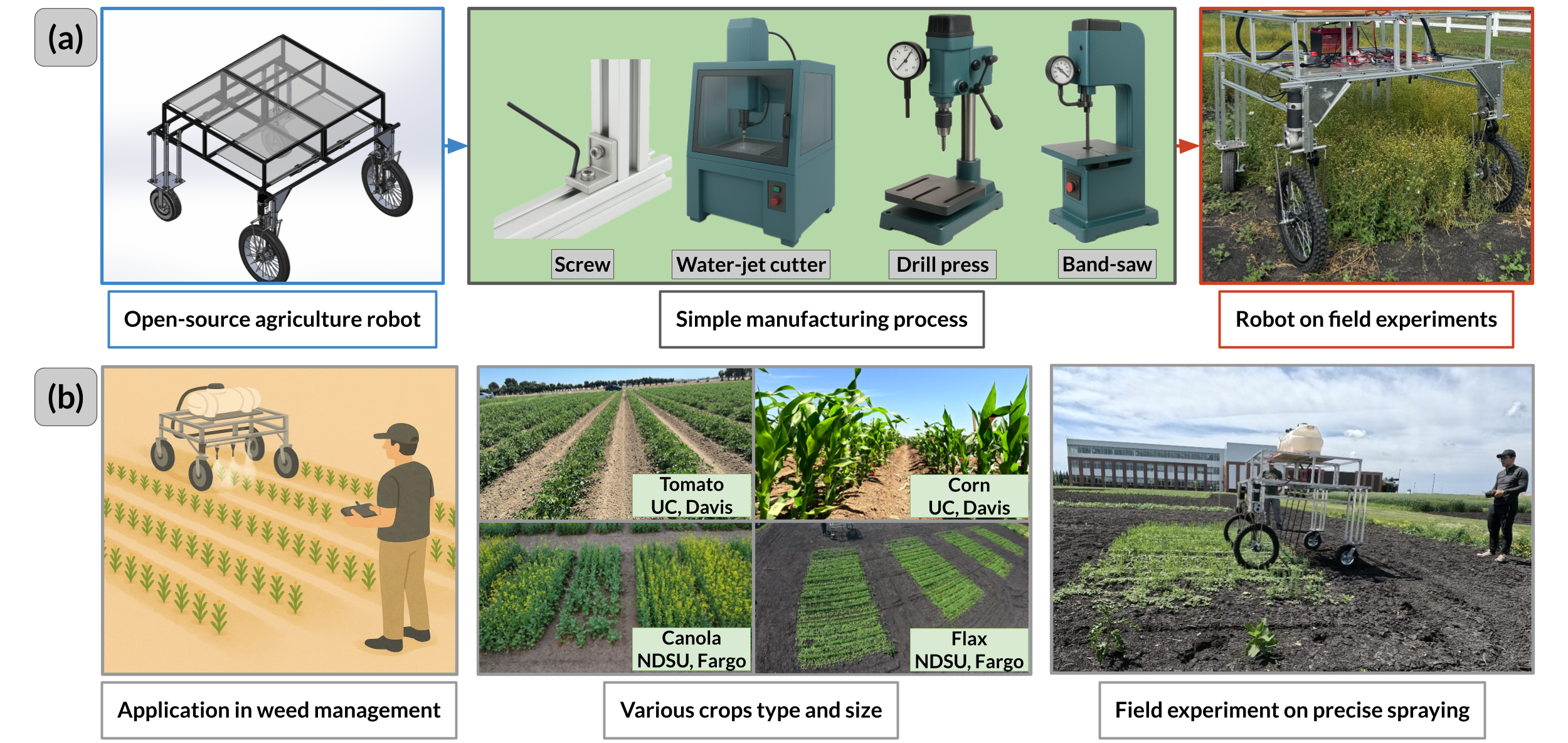}
    \caption{Overview of the AgriCruiser: (a) 3D model of the open-source agricultural robot, designed with simple manufacturing processes using accessible machining, and performance validated through field experiments. (b) Application in weed management, where the AgriCruiser was evaluated on various crop types and sizes, successfully performing precise spraying for weed control.}
    \label{fig: overview}
\end{figure}

The main contributions of this paper are summarized as follows:
\begin{itemize}
  \item A \textbf{modular over-the-row chassis} design with adjustable track width and high clearance capable of adapting to different crop types and field layouts.
  \item The development of a \textbf{large agile platform} with sufficient payload capacity to accommodate additional subsystems for various agricultural applications and \textbf{compact maneuvering} via differential drive, enabling 0.79 m pivot‑turn radii for tight headlands.
  \item \textbf{Precision sprayer integration}, capable of herbicide application and data collection to support efficient weed management.
  \item \textbf{Field validation} demonstrating large reductions in weed density with lower crop damage, plus cross‑terrain and multi-crops traversal consistent with torque sizing.
\end{itemize}
Beyond immediate weed management, the AgriCruiser provides a reusable experimental platform for agricultural robotics: its open designs, low cost, and reconfigurability make it suitable for rapid iteration on sensing, actuation, and autonomy. The design and concept is ideal for tasks such as crop phenotyping, irrigation, soil probing, image data collection, and even fruit picking. This adaptability makes our robot an ideal solution for farms looking to integrate technology across multiple stages of the planting process \cite{different_application}. The rest of the paper is organized as follows: Section~\ref{sec: related-works} reviews related work in modular ground robots and plot‑scale weeding systems, Section~\ref{sec: design} details the mechanical design and analysis, Section~\ref{sec: electronics} presents the electronics and software architecture, Section~\ref{sec: experiments} describes the experiments conducted to evaluate the design and its results, and finally, Section~\ref{sec: conclusion} concludes the study with future work.

\section{Related Works}\label{sec: related-works}

In recent years, a variety of agricultural robots have been researched and developed, each system targeting specific tasks or crop types \cite{agribot_review_2}. This section examines current developments in agricultural robot platforms, focusing on modular ground systems and precision weeding technologies.

\subsection{Commercial Agricultural Robot Platforms}
Commercial agricultural robots can be categorized into task-specific and multi-purpose platforms. Task-specific systems target individual applications but lack adaptability across diverse farming operations. In 2024, MQ Autonomous Agritech introduced the M200 Autonomous Sprayer, offering a safe and autonomous solution for weed spraying. The system is tailored for orchard and vineyard farms, with functionality restricted to a specific spraying task \cite{m200}. In the same year, Odd-Bot also introduced Maverick, which provides another autonomous solution for weed control. The robot features a mechanical weed-uprooting system and is primarily designed for low-organic crops, such as carrots. Due to the low ground clearance resulting from the uprooting system located below the robot, the system cannot traverse over tall crops, such as corn \cite{maverick}. If we see commercial harvesting robots, the Tortuga Harvesting Robot is recognized for its advances in the precision harvesting of strawberries and grapes \cite{agtecher}. In addition, Agrobot is another commercial robotic strawberry harvester \cite{agrobot}. With that being said, these systems specialize in a specific task, which is fruit harvesting, and are only designed for certain particular crops.

Multi-purpose commercial platforms offer greater versatility but come with significant cost barriers. The Thorvald platform (SAGA Robotics) features modular reconfiguration capabilities with implementations costing \$60,000-\$100,000. While Thorvald 3 has shown 60-90\% pesticide reduction in vineyard applications, its high cost limits adoption among small-to-medium farms. The Weed Terminator, based on Thorvald technology, operates with fixed chassis dimensions. Despite advertised customizability, such customization is limited to the ordering stage, limiting its adaptability to various crop layouts \cite{thorvald_paper}. Robotics Plus launched Prospr, an autonomous and multi-use vehicle designed to carry out a variety of crop tasks more efficiently. However, the robot requires a minimum row spacing of at least 6 feet (1.83 m), which poses challenges for operation on certain farms due to its considerable size \cite{prospr}. 

Recent commercial developments include Farm-NG's Amiga platform, which provides modular electric propulsion with four-wheel hub motors and costs significantly less than traditional alternatives \cite{farmng}. Although it is designed for ``between rows or above crops," it cannot effectively traverse over tall crops during their mature growth stages, limiting its operational window throughout the growing season.


\subsection{Research \& Open-Source Platforms}
Academic research has produced several notable platforms addressing cost and accessibility challenges. Open-source initiatives like FarmBot \cite{farmbot-genesis} provide CNC-style precision agriculture for small-scale applications but lack the mobility and clearance required for field-scale operations. The Hefty platform, built on Farm-NG's Amiga base, exemplifies modular design principles for agricultural manipulation research, demonstrating reconfigurability across sensing, computing, and mounting systems \cite{guri2024hefty}. Reconfigurable research platforms have shown promise for multi-crop applications. The Reconfigurable Ground Vehicle (RGV) developed for corn navigation features shape-changing capability between parallel and linear configurations, enabling both inter-row and intra-row traversal \cite{schmitz2022design}. Although due to the smaller size of the robot the application is limited to scouting and data collection.

In conclusion, most existing agricultural robots face a standard limitation: these systems lack either the adjustability needed to handle various crop types, sizes, and row layouts or the adaptability required to integrate additional subsystems for multiple tasks. These drawbacks restrict their broader adoption and highlight the need for a more versatile platform, which is addressed in the design of the AgriCruiser. The AgriCruiser addresses these limitations through three key design principles: reconfigurability, affordability, and agility. Unlike existing systems that are highly specialized or prohibitively expensive, the AgriCruiser provides dynamic track width adjustment (1.42-1.57 m) using a T-slot framing system, enabling rapid field reconfiguration without disassembly. With 0.94 m of ground clearance and in-place rotation capability, it operates across diverse crops (flax, canola, wheat) from early growth through harvest stage. Most importantly, as a fully open-source platform, it enables complete customization and replication, eliminating vendor lock-in while supporting multiple agricultural tasks including weed management, spraying, phenotyping, and potentially harvesting.

\subsection{Precision Weed Management Systems}

In agricultural robotics, weed management is one of the most studied topics, second only to harvesting \cite{agribot_review}. The number of research publications focusing on weed control, whether to improve precise spraying or alternative approaches, reflects its economic importance. These studies also often highlight concerns about the environment and human health, particularly the association of exposure to pesticides. Although mechanical approaches such as uprooting or trimming can reduce the use of pesticides, spraying often remains more cost-effective and time-efficient. Therefore, precise spraying and safe application of pesticides have been the subject of intensive research in the past decade, exceeding many mechanical approaches to weed management \cite{weed_management}.

Aerial versus terrestrial systems present distinct trade-offs. While aerial platforms like DJI's AGRAS MG-1S \cite{agras} offer field coverage, they have many disadvantages, including the risk of environmental contamination due to mid-air spraying, inconsistent pesticide coverage, and high sensitivity to weather conditions. Another limitation of aerial spraying robots is their low payload capacity. For example, the AGRAS MG-1S has an operating payload of only 10 kg, which restricts the amount of pesticide carried, thus requiring more refilling when covering large fields. Terrestrial systems provide better precision and payload capacity but require robust navigation systems for autonomous operation.

For terrestrial spraying, robots also vary widely in design and size. An example is a compact, low-cost robot that can fit between rows of crops \cite{mini_robot}. Although the robot was able to move on the soil terrain and completed spraying experiments, its ability to operate on more extreme, uneven terrains has not been demonstrated. Due to its small size, it may not have enough power to overcome uneven terrains and other challenging surfaces commonly found at farm sites. Moreover, its small size makes it inefficient for covering large fields, reducing its suitability for large-scale operations. Another terrestrial spraying robot is the AVO robot sprayer, developed by Ecorobotix. The robot has been proven to overcome obstacles, operating on slopes up to 10°, and provides high maneuverability with a small turning radius. However, its total weight is 750 kg, including batteries and a 120-liter herbicide tank filled, which is heavier than a compact tractor \cite{beckside}. As mentioned above, such a heavy system can contribute to soil compaction and increase the risk of crop damage.

Thus, the AgriCruiser demonstrates precise spraying for efficient weed management, with the ability to carry a high payload and traverse farm sites reliably without causing soil compaction or damaging surrounding crops. Field validation demonstrates exceptional weed control effectiveness. Additionally, the AgriCruiser is capable of continuously spraying for up to 30 minutes, covering up to 1003.35 square meters of crop area, making it an effective platform for weed management.

\section{Mechanical Design}
\label{sec: design}
\subsection{Structure Design}
\label{subsec:structure_design}
The primary objective is to develop a reconfigurable chassis that adjusts to diverse crop sizes and field requirements. The T-slot framing system enables rapid assembly and modification without permanent connections, addressing the critical need for adaptability in agricultural robotics. The T-slot extrusion system provides several advantages over traditional welded chassis designs. Mechanical fastening rather than welding maintains reconfigurability, as joints can be loosened and repositioned as needed. Components mounted on T-slot extrusions can be repositioned within millimeters and securely fixed using hex keys, enabling precise field adjustments. The system utilizes aluminum extrusions offering optimal strength-to-weight ratio for agricultural applications. Standard cutting and assembly tools eliminate the need for specialized machinery, supporting the platform's open-source accessibility. A comprehensive range of off-the-shelf components—including angle brackets, gussets, and hinges—ensures worldwide availability and cost-effectiveness. As this is an open-source platform, accessibility to materials and components while ensuring the robot operates seamlessly is of top priority.

\begin{figure}[h!]
    \centering
    \includegraphics[width=\linewidth]{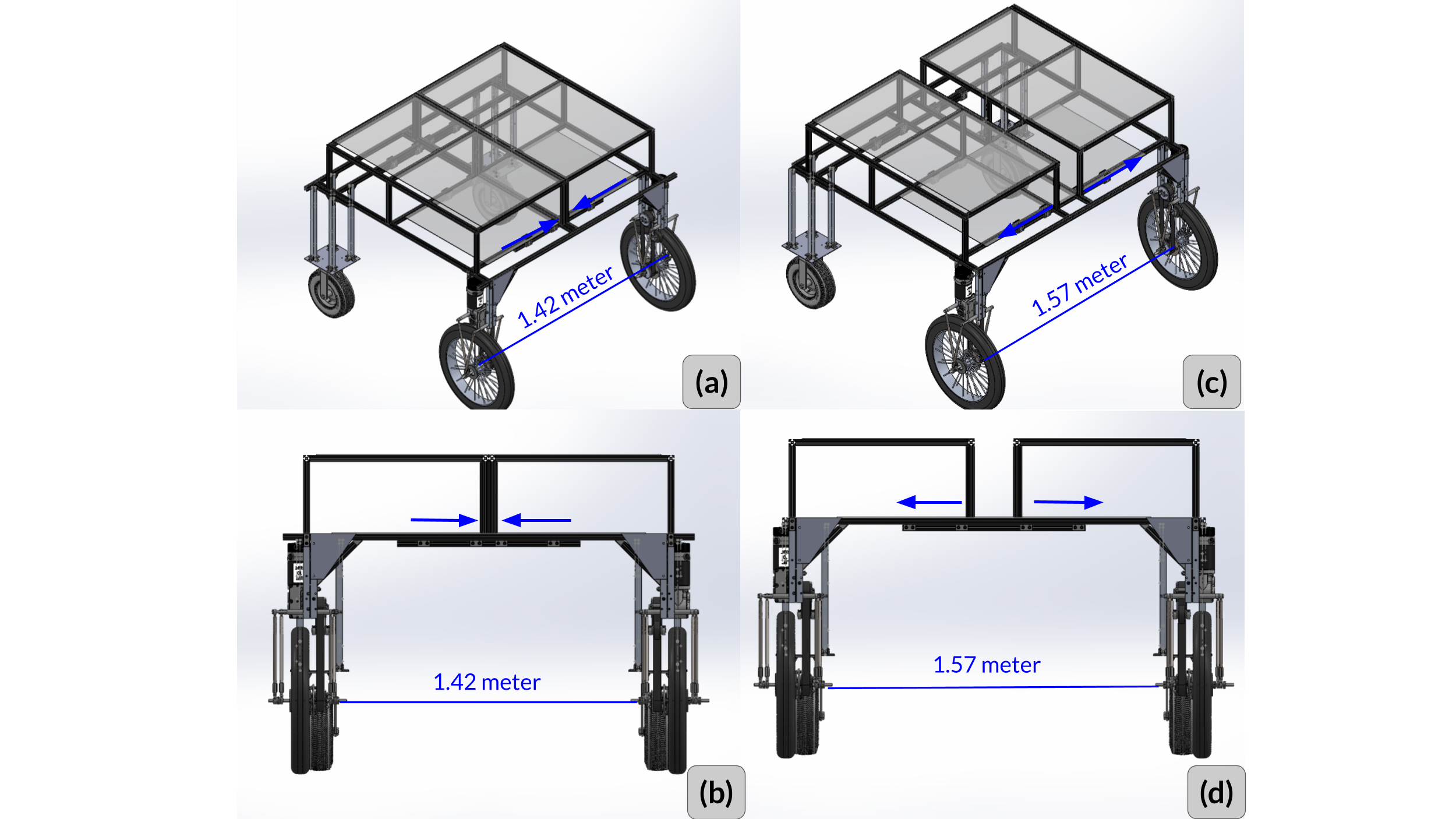}
    \caption{Robot at its (a) fully closed track width from top view, (b) fully closed track width from side view, (c) fully extended track width from top view, and (d) fully extended track width from side view }
    \label{fig: Robot_Width}
\end{figure}

The chassis provides dynamic track width adjustment from 1.42 m to 1.57 m (measured between driven wheel centers), accommodating diverse crop layouts without disassembly. The minimum width is 1.42 m (Figure~\ref{fig: Robot_Width}a - ~\ref{fig: Robot_Width}b), while the maximum width is 1.57 m. (Figure~\ref{fig: Robot_Width}c - ~\ref{fig: Robot_Width}d). This 0.15 m adjustment range enables operation across multiple crop types (see Table~\ref{tab: Trackwidth}) while maintaining structural integrity.

\begin{figure}[h!]
    \centering
    \includegraphics[width=\linewidth]{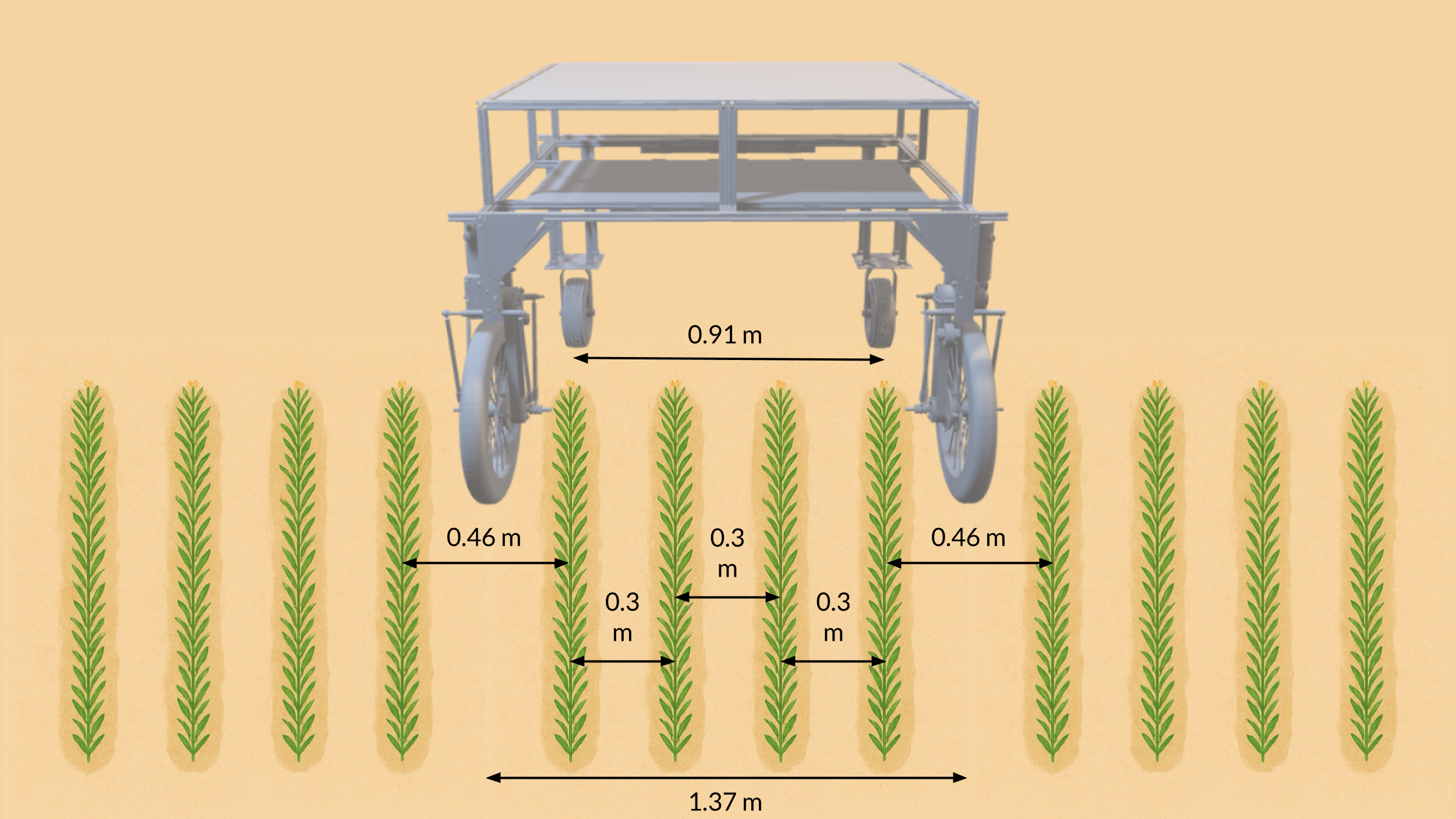}
    \caption{Rendering of the robot in a flax field, illustrating that the track width covers the crop width while the wheel width fits within the row spacing}
    \label{fig: Robot_Field_Render}
\end{figure}

Operational clearance between the wheels and the crop rows are critical to avoid damage, in case the robot deviates from its straight path. Given that the maximum overall width of the robot is approximately 1.57 m, plots up to 1.52 m in width can be accommodated between the wheels of the robot. Field validation (see Section~\ref{sec: experiments}) in flax crops demonstrates effective operation: four-row plots spanning 1.37 m were successfully traversed with the robot configured at an appropriate track width (Figure~\ref{fig: Robot_Field_Render}), completing its designated tasks without damaging the crop. 

Another important consideration is the row spacing, as this determines the actual wheel path. The spacing must be larger than the wheel width to prevent crop damage. In the flax field experiment mentioned above, the row-to-row distance was 0.46 m (Figure~\ref{fig: Robot_Field_Render}). Since the width of the wheels is slightly less than 0.36 m, this provides a clearance of 0.05 m on each side, ensuring that the wheels remain a safe distance from the crop, even in cases of deviations.

Table~\ref{tab: Trackwidth} and Table~\ref{tab: Rowspacing} show the crop dimensions collected in collaboration with the faculty of the institutions in the respective areas. The AgriCruiser has already been successfully deployed on the flax and canola fields in Fargo, North Dakota, confirming its suitability for these crops. Although field experiments were not conducted in Davis, California, for tomatoes or corn, the information on crop dimensions was taken into our design considerations and we are confident that the AgriCruiser is capable of operating on these fields too. 

\begin{table}[h!]
\centering
\begin{minipage}{0.40\linewidth}
    \centering
    \begin{tabular}{|c|c|c|}
        \hline
        \textbf{Region} & \textbf{Crop type} & \textbf{Track width} \\
        \hline Fargo & Canola & 1.37 m \\
        \hline Fargo & Flax & 1.42 m \\
        \hline Davis & Tomato & 1.52 m \\
        \hline Davis & Corn & 1.52 m \\
        \hline
    \end{tabular}
    \caption{Track width dimension}
    \label{tab: Trackwidth}
\end{minipage}
\hspace{1cm}
\begin{minipage}{0.40\linewidth}
    \centering
    \begin{tabular}{|c|c|c|}
        \hline
        \textbf{Region} & \textbf{Crop type} & \textbf{Row-to-Row Spacing} \\
        \hline Fargo & Canola & 0.41 m \\
        \hline Fargo & Flax & 0.46 m \\
        \hline
    \end{tabular}
    \caption{Row-to-Row Spacing dimension}
    \label{tab: Rowspacing}
\end{minipage}
\end{table}

The robot has an overall height of approximately 0.94 m, allowing it to traverse above crops shorter than this threshold. According to Flax Production in North Dakota \cite{flax_production}, flax typically grows to a height of 24-36 in. (0.61-0.91 m). This indicates that the robot is capable of traversing flax crops, even at full growth. In the experiments reported in Section~\ref{sec: experiments}, the focus was on weed management; therefore, weed spraying was performed during the early growth stage, when the crop was still relatively low. The robot was also brought back to the same flax field two months after the initial spraying, to demonstrate its ability to traverse over the flax during the maturation stage of its growth cycle with increased crop height. Weed spraying is prohibited during the seed maturation stage, as herbicides on flaxseed would pose risks to food safety. However, if redesigned for different applications, such as harvesting or phenotyping, the robot can still operate effectively as long as the crop height remains below 0.94 m. 

\subsection{Manufacturing Process}
\label{subsec: manufacturing_process}
\begin{figure}[h!]
    \centering
    \includegraphics[width=\linewidth]{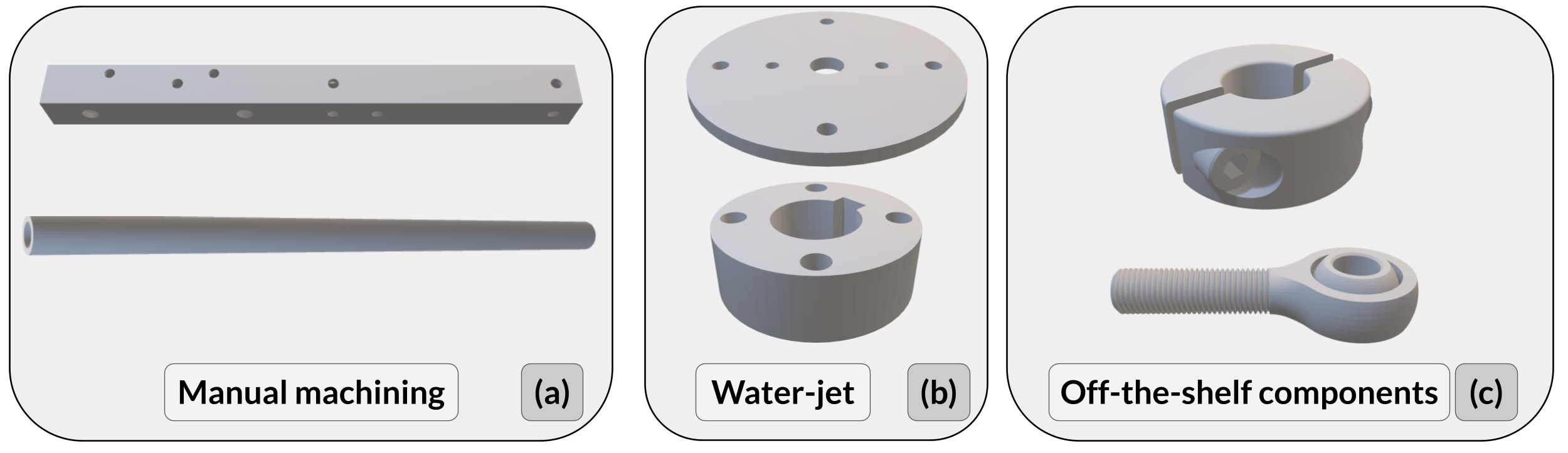}
    \caption{3D models of components during the manufacturing process: (a) parts produced by manual machining, such as drilled tubes (b) parts fabricated using a water-jet cutter, such as adapters and (c) off-the-shelf parts, such as ball joints and shaft collars.}
    \label{fig: manufacturing}
\end{figure}

As mentioned, the AgriCruiser is designed as an open-source platform, with one of the main priorities being the provision of detailed guidelines with simple manufacturing processes using widely accessible machinery. Most components can be manufactured with basic manual machines such as drill presses, band saws, and water-jet cutters. Although the use of more advanced machines, such as milling or CNC machines, can further simplify the process, they are not strictly required. For components requiring more advanced machines, a variety of companies provide custom fabrication services at affordable prices, offering options such as water-jet cutting, laser cutting, and CNC machining. Since all components have already been designed in SolidWorks, fabrication services typically require only the export of files in standard formats, such as STEP or DXF. Figure~\ref{fig: manufacturing} shows the 3D models of some selected custom components, each manufactured by water-jet cutting or manual machining. 

The remaining components consist of various available off-the-shelf products. A detailed bill of materials is provided to ensure accessibility, and most components are available from suppliers such as McMaster-Carr, Amazon, or local hardware stores. Table~\ref{tab: manufacturing_process} summarizes the breakdown of the manufacturing processes used for robot components.

\begin{table}[h!]
    \centering
    \begin{tabular}{|c|c|c|}
        \hline
        \textbf{Methods} & \textbf{Number of Components} & \textbf{Percentage of Total} \\
        \hline Water-jet & 22 & 7.05\% \\
        \hline Manual machining & 24 & 7.69\% \\
        \hline Off-the-shelf components & 266 & 85.26\% \\
        \hline
    \end{tabular}
    \caption{Breakdown of manufacturing processes for the robot transmission system, which consists of 312 components in total}
    \label{tab: manufacturing_process}
\end{table}

As shown in Table~\ref{tab: manufacturing_process}, only 14.74\% of the total system require custom fabrication, out of which just 7.05\% of the components were manufactured through an outsourced fabrication company, 7.69\% were manufactured in-house using manual machining such as a bandsaw or drill press, and most are off-the-shelf components. The platform emphasizes the use of custom parts and standardized components to simplify the manufacturing process and maximize worldwide accessibility, thereby supporting the open-source mission.

The chassis is constructed from approximately 25.91 m of T-slot extrusion, which is cut into smaller segments for assembly. Extrusions can be bought in various lengths, as provided in the bill of materials, and longer extrusions can be cut to the required lengths using a bandsaw. The segments are joined together using L-brackets or side mounts, allowing the chassis to be assembled by simply cutting to size and fastening with screws. Hence, the platform achieves global replicability through minimal custom fabrication, standardized components, and basic tooling requirements.

\subsection{Differential-Drive Kinematic Model}
\label{subsec:differential_drive_kinematic_model}
The robot features a front wheel differential drive system, consisting of two motorized wheels at the front and two passive caster wheels at the rear.  The key advantage of a differential drive is that each wheel is powered by an independent motor, allowing separate speed control, creating diverse motion patterns without requiring complex steering mechanisms. When both wheels rotate at identical speeds in the same direction, the robot achieves straight-line motion. Pivot turning occurs when one wheel remains stationary while the other rotates, creating rotation around the fixed wheel. In-place rotation results from both wheels rotating at equal speeds in opposite directions, enabling zero-radius turns crucial for headland maneuvers. Both turning capabilities offer a minimal turning radius which is vital in agricultural settings, where crops are typically planted in dense spacing to maximize land use efficiency. Once the robot completes a task in one row, it is designed to make a U-turn and proceed into the next row to continue the operation. If the turning radius were large, the robot might damage the surrounding plots. These motion characteristics can be analyzed through kinematic modeling, which allows evaluation of robot motion. 

In a differential-drive system, the driven wheels define the motion, while the casters are passive wheels that roll and turn to accommodate the trajectory imposed by the driven wheels. Due to the symmetry and uniform dimensions of the driven wheels, the velocities of both the driven wheel and the caster wheel can be combined into one variable, denoted $v_r$ and $v_l$. Consequently, the linear velocity of the robot, $V$, and the angular velocity, $\omega$, are determined from $v_r$, $v_l$ and the track width L \cite{kinematic_model}. 
\begin{equation}
V = \frac{v_r + v_l}{2}, \qquad 
\omega = \frac{v_r - v_l}{L}.
\end{equation}

The general motion of the robot can be described in terms of its $X$, $Y$, and $\theta$ (heading angle), as determined by the velocity and orientation of the vehicle. 
\begin{equation}
\dot{X} = V \cos\theta, \qquad
\dot{Y} = V \sin\theta, \qquad
\dot{\theta} = \omega.
\end{equation}

The turning radius $R$ is defined as the distance from the midpoint of the wheel axle to the instantaneous center of rotation, which can be expressed as:
\begin{equation}
R = \frac{V}{\omega} = \frac{L}{2} \cdot \frac{v_r + v_l}{v_r - v_l}.
\end{equation}

For in-place turning, the wheel velocities are $v_r = -v_l$. Substituting into the above turning radius equation yields $R = 0$, indicating that the center of the chassis rotates about its own position. However in our case the center is shifted to the center of the two driven wheels. For pivot turning, one wheel remains stationary while the other turns. For example, if the right wheel is fixed ($v_r = 0$), substitution into the same equation above gives $R = \frac{L}{2}$, which means that the robot center follows a circular path of radius $L/2$.

These equations establish a foundation framework for analyzing and understanding the kinematics of a four-wheeled robot with two-wheel differential drive under various operating conditions and different scenarios.

\subsection{Transmission System}
\label{subsec:transmission_system}
In agricultural settings, the transmission system is a critical subassembly that ensures the robot can traverse uneven terrain. Once deployed in the field, the robot must adapt to various surface conditions such as concrete, grass, gravel, or soil. In our first iteration, the AgriCruiser featured four motorized wheels driven by PMDC brush motors. In this old design, the robot ran smoothly on paved roads, but its performance on soil was unsatisfactory. The problem arose from the fact that soil has high deformability, uneven surface profiles, and greater rolling resistance due to sinkage and localized inclines. These factors make locomotion significantly more difficult compared to flat and rigid concrete. As a result, the first iteration of the AgriCruiser failed to achieve mobility on the fields.

\begin{figure}[h!]
    \centering
    \includegraphics[width=\linewidth]{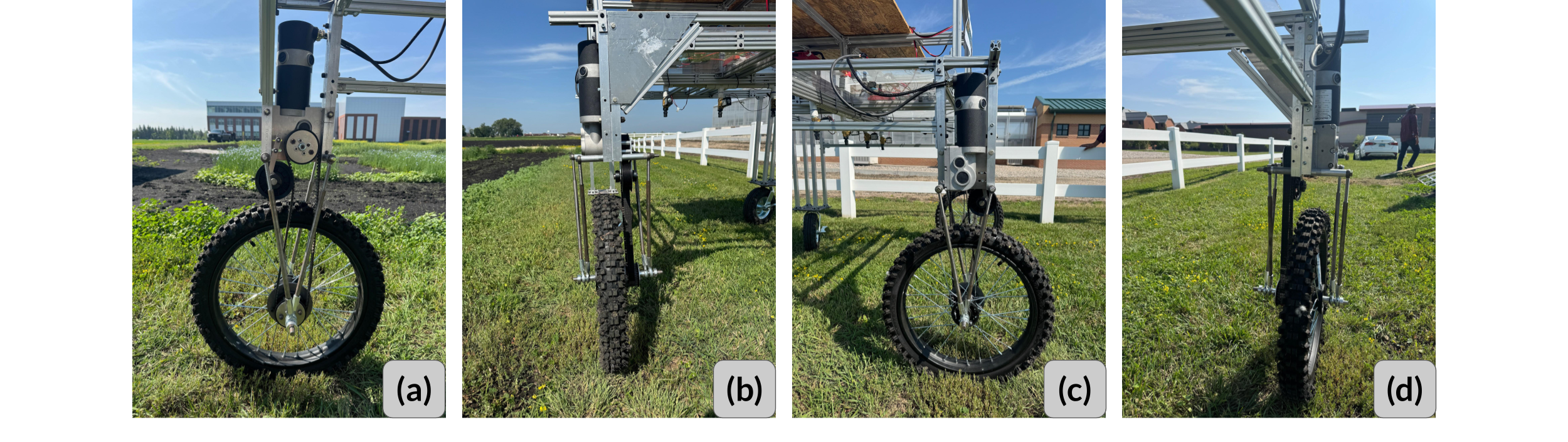}
    \caption{Transmission system equipped with ElectroCraft MP26 motors shown in a \ang{360} view: (a) inner side view, (b) front view, (c) outer side view, and (d) rear view}
    \label{fig: transmission_system}
\end{figure}

To address these limitations, the transmission system was redesigned to provide higher torque for challenging agricultural settings. The new design features a two-wheel front-drive system, powered by two low-RPM, high-torque PMDC brush motors. The selected motor is the MP26 from ElectroCraft, configured with a gear ratio of 32:1, which provides continuous torque up to 31.42 Nm for 15 minutes (Figure~\ref{fig: transmission_system}). The two-wheel drive was selected over the four-wheel drive to reduce cost and mechanical complexity while still providing sufficient torque for field conditions. Motor selection was determined by torque requirement calculations, as detailed below. The parameters and assumptions for the calculation were defined in a worst-case scenario, including a high-velocity requirement, a rolling resistance coefficient of up to 0.2, and localized inclines of \ang{10}. For comparison, soft soil typically has a rolling resistance coefficient in the range of 0.04 to 0.08 (Table~\ref{tab: rrc}), indicating that the assumptions ensure reliable performance across a variety of soil types. 

\begin{table}[h!]
    \centering
    \begin{tabular}{|c|c|}
        \hline
        \textbf{Surface Conditions} & \textbf{Rolling Resistance Coefficient} \\
        \hline Concrete & 0.002  \\
        \hline Asphalt & 0.004  \\
        \hline Rough Paved Road & 0.008  \\
        \hline Gravel & 0.02  \\
        \hline Soil (Medium-Hard) & 0.04-0.08  \\
        \hline Sand & 0.2-0.4 \\
        \hline
    \end{tabular}
    \caption{Rolling Resistance Coefficient \cite{engineeringtoolbox}}
    \label{tab: rrc}
\end{table}

In addition to variation across different terrains, significant variability also exists within a single field. For example, entry paths are often compacted, while crop rows may contain softer soil, and moisture differences can further alter traction. This variability requires a transmission system with sufficient torque to accommodate sudden changes in field conditions \cite{terrain}.

Based on the parameters and assumptions above, the governing equation is derived from a free-body force balance of a wheel, where the tractive force of the wheels is the sum of the inertial force, the gravitational component and the rolling resistance, as given by Equation~\ref{eq: Tractive_force}
\begin{equation}
    F_\text{t} = F_\text{i} + F_\text{g} + F_\text{rr}
\end{equation}
\begin{equation}
    F_\text{tractive} = m a + m g \sin\theta + F_\text{rr},
    \label{eq: Tractive_force}
\end{equation}
where $m$ is the vehicle mass, $a$ is the linear acceleration, $g$ is the gravitational acceleration, $\theta$ is the incline angle, and $F_r$ is the rolling resistance force.

Rolling resistance is modeled as
\begin{equation}
    F_\text{rr} = C_{rr} \, m g \cos\theta,
\end{equation}

where $C_{rr}$ is the rolling resistance coefficient, which can vary with terrain. Substituting this expression, the required tractive force can be rewritten as 
\begin{equation}
    F = m a + m g \sin\theta + C_{rr} \, m g \cos\theta,
\end{equation}

The tractive force in the tire contact patch is related to the torque of the wheels by Equation~\ref{eq: Wheel_force_torque}, where $r$ is the radius of the wheel. For a system with $n$ driven wheels, the torque per motor is equal to the total required torque divided by the $n$ wheels.
\begin{equation}
    \tau_\text{wheel} = F \, r \qquad
    \tau_\text{motor} = \frac{\tau_\text{wheel}}{n}
    \label{eq: Wheel_force_torque}
\end{equation}

Finally, wheel speed relates linear velocity to angular velocity, which can be further expressed in revolutions per minute (RPM) for the motor specification as Equation~\ref{eq: RPM}. 
\begin{equation}
    \omega = \frac{v}{r} \qquad
    \text{RPM} = \frac{v \cdot 60}{2 \pi r}.
    \label{eq: RPM}
\end{equation}

For a mobile vehicle with front wheel drive and rear caster, the two front wheels generate motion while the rear caster wheels passively follow. The robot has a track width of $2b$, a wheelbase of $L$, a center of gravity height of $h_{cg}$, and a yaw inertia of $I_z$. Let $F_L$ and $F_R$ be the longitudinal forces applied by the front wheels, and let a rear caster be at position $\mathbf{r}_c = (x_c, y_c)$, with $x_c = -L$ and $y_c = \pm b$. 

The linear and angular accelerations of the center of mass are given, respectively, by:

\begin{equation}
    F_{L} = F_{R} = \frac{F}{2}, \qquad \qquad
    a_x = \frac{F_L + F_R}{m}, \qquad \qquad
    \alpha_z = \frac{(F_R - F_L) \cdot b}{I_z}
    \label{eq: accel_COM}
\end{equation}

The velocity of each caster wheel is determined as follows:
\begin{equation}
    \mathbf{v}_c = \mathbf{v}_{cm} + \boldsymbol{\omega} \times \mathbf{r}_c
    = 
    \begin{bmatrix} a_x - \alpha_z y_c \\ \alpha_z x_c \end{bmatrix}
\label{eq: caster_vel}
\end{equation}

Since caster wheels are passive, the forces and torques they experience arise from \textit{ground reaction} and \textit{bearing friction}, rather than from direct acceleration of a portion of the robot mass. The vertical load on each caster is:
\begin{equation}
    N_{CL/CR} = \left( \frac{L_f}{L} \cdot mg \right) \pm \frac{m a_x h_{cg}}{2L}
    \label{eq: vertical_load}
\end{equation}

The rolling torque required for caster wheels to overcome wheel friction is the following:
\begin{equation}
    \tau_{\text{roll}} = N \cdot C_{rr} \cdot r
    \label{eq: roll_torque}
\end{equation}

The swivel torque needed to align the caster with the velocity vector $\mathbf{v}_c$ is:
\begin{equation}
    \tau_{\text{swivel}} = \tau_{\text{sf}} + d \, \omega_{\text{swivel}},
    \label{eq: swivel_torque}
\end{equation}
where $\tau_{\text{sf}}$ is the static friction torque, $d$ is the viscous friction coefficient and $\omega_{\text{swivel}}$ is the swivel rate.

The governing equations are listed above, particularly Equation~\ref{eq: Wheel_force_torque}, which represents the derived expression that incorporates all relevant relationships and guided the selection of motors and wheel dimensions for driven wheels. To the best of our knowledge, the AgriCruiser introduces a novel configuration in agricultural robotics at this scale by strategically combining two-wheel front differential drive with two passive rear caster wheels. This hybrid approach achieves exceptional maneuverability advantages while simultaneously optimizing energy efficiency through the elimination of wheel skidding that commonly occurs in four-wheel drive systems during turning maneuvers. The supporting equations for both the driven and the caster wheels were also applied to verify the ability of the system to traverse uneven terrain, thus ensuring mobility before field deployment.

\subsection{Herbicide Sprayer System}
\begin{figure}[h!]
    \centering
    \includegraphics[width=\linewidth]{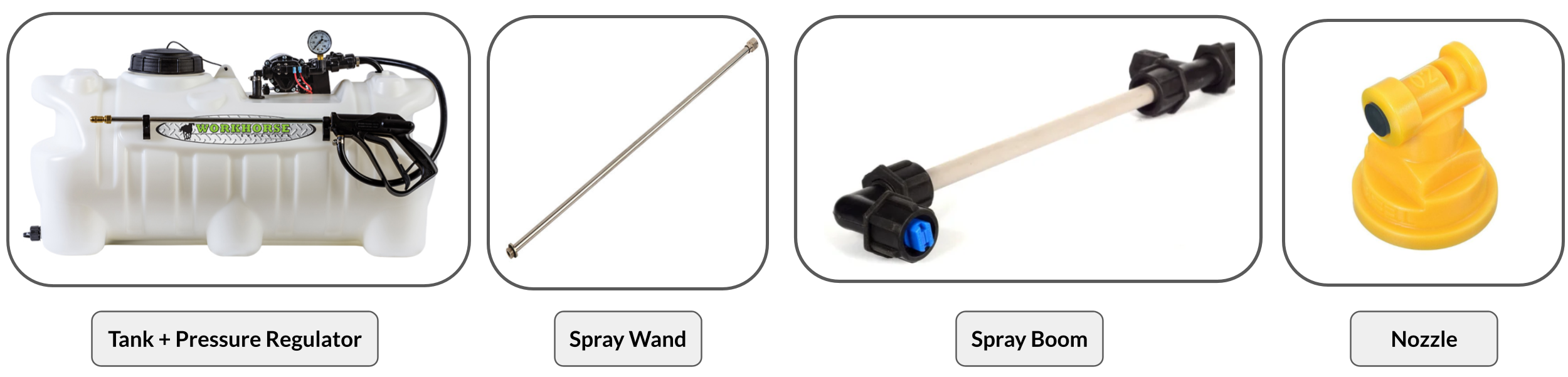}
    \caption{Sprayer system components, including the herbicide tank with pressure regulator attached on top, the solenoids attached to the sprayer boom and wand for herbicide delivery, and the spray nozzles}
    \label{fig: sprayer_system}
\end{figure}

The integrated herbicide sprayer system demonstrates the AgriCruiser's modular subsystem integration capability while providing precise weed management functionality. The system prioritizes operational efficiency, chemical precision, and field coverage optimization through carefully selected components and validated operational parameters. It comprises six primary components: the herbicide tank, pump, bypass valve for pressure regulation, solenoid valve for open/close control, spray boom, and wand equipped with nozzles. In this situation, the bypass valve maintains operating pressure by recirculating excess flow back to the tank, as shown in Figure~\ref{fig: sprayer_system}.

The herbicide tank has a capacity of 25 gallons (94.64 L). The pressure regulator includes a knob to adjust and maintain the desired operating pressure, which controls herbicide flow during spraying. In the weed management experiments, the operating pressure was maintained at approximately 40 PSI. According to the nozzle datasheet, the flow rate at this pressure was 0.2 gallons per minute (GPM) for each nozzle. The selected operating pressure was considered optimal because higher pressures can increase flow rates and finer misting, which leads to excessive herbicide usage.

Four nozzles were attached to the sprayer system, resulting in a total flow rate of 0.8 GPM at a constant operating pressure of 40 PSI. During the flax rows experiment, it took approximately 4 seconds to spray one flax plot. Twelve plots were sprayed in total, corresponding to a total spraying time of approximately 48 seconds and a consumption of approximately 0.65 gallons (2.46 L) of herbicide.

When fully filled to its 94.64 L capacity, which still remains within the payload capacity, the system can spray continuously for approximately 30 minutes. Since one plot covers an area of 2.23 square meters and it takes about 4 seconds to spray each plot, the AgriCruiser can cover up to 1003.35 square meters of crop area under the same operating conditions.

\subsection{Specifications}
The platform itself weighs approximately 20 kg. Combined with additional components, such as herbicide tank, battery, sprayer system, and minor hardware, together weighing approximately 25 kg. The total payload experienced by the transmission system was approximately 45 kg during the experiments. Under average terrain conditions, between saturated and unsaturated soils, the calculated velocity is approximately 5.7 m/s, as derived from the governing equation provided in Subsection~\ref{subsec:transmission_system}. However, for safe operations the driving motors were set to operate only 15\% of their full capacity, the theoretical peak velocity under these conditions was about 0.85 m/s.

During field experiments, the AgriCruiser was tested on unsaturated soil, covering an 8-foot (2.44 m) crop row in 4 second, corresponding to a velocity of approximately 0.61 m/s. The experimental velocity is reasonably close to the theoretical value, considering that in realistic agricultural conditions, numerous factors such as traction loss due to muddy patches or extremely uneven terrain, affect overall performance beyond the idealized assumptions made during calculation.

Furthermore, assuming that the motors are operated at their full capacity and maintaining an experimental velocity of 0.61 m/s, the maximum payload that the AgriCruiser can carry is estimated at 100 kg with a factor of safety of 1.5. The payload does not include the weed management system, thus estimating the maximum payload capacity of the platform when repurposed for agriculture applications beyond weed management.

\begin{figure}[h!]
    \centering
    \includegraphics[width=\linewidth]{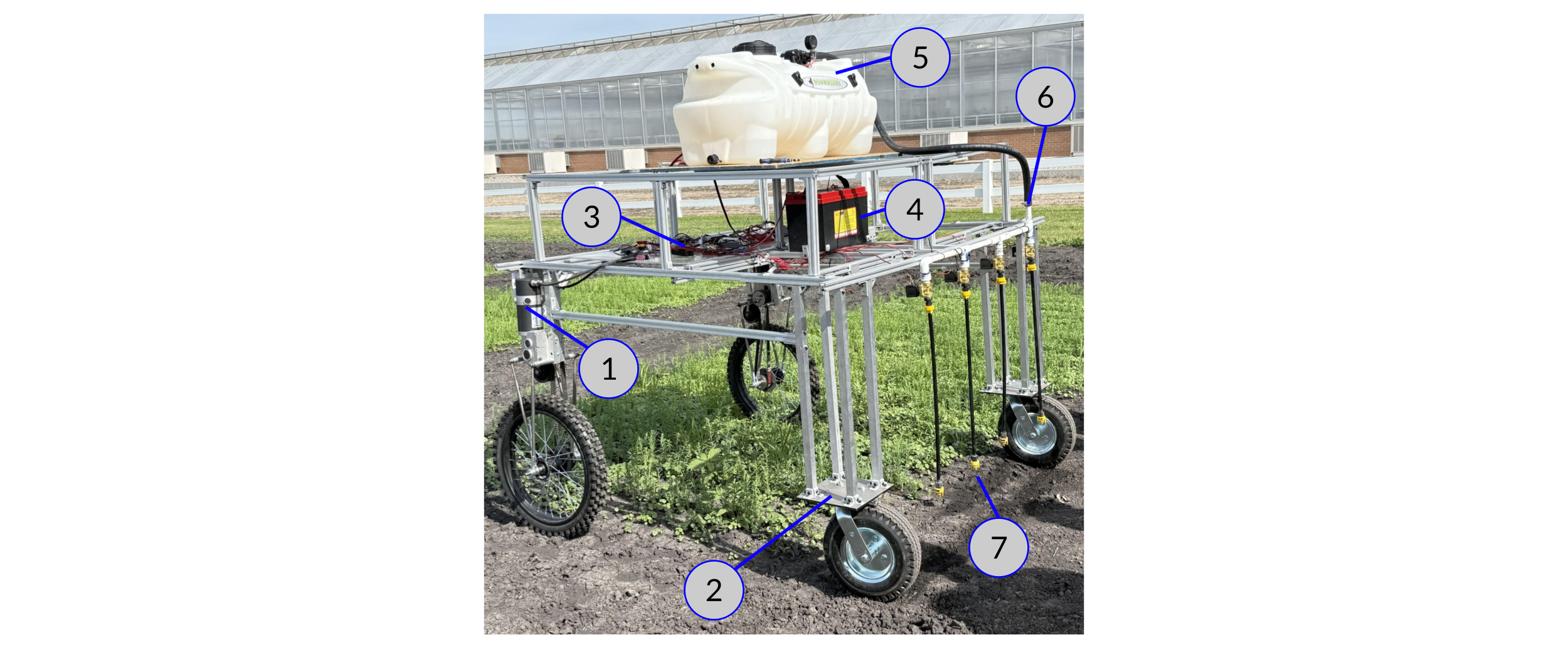}
    \caption{Main components of the AgriCruiser: (1) transmission system of the driven wheels; (2) rear caster wheel; (3) electronics and hardware; (4) battery; (5) herbicide tank with pressure regulator mounted on top; (6) solenoids attached to the sprayer boom; (7) sprayer wand with spraying nozzle}
    \label{fig: major_component}
\end{figure}

\section{Electronics System \& Software Architecture}
\label{sec: electronics}

\subsection{Electronics System}

This section focuses on the electronics system and its layout for the robot. The goal of the electronics system is to control and distribute electricity throughout the machine. The system is structured in a modular way so that each part can be easily identified and isolated if issues arise. The operations on the AgriCruiser is controlled through a Radiomaster controller, enabling remote driving and activation of the sprayers. To accomplish this, the electronics system integrates several components that work together to provide power, control, and communication. The following paragraphs explains the individual component utilized in the project.

The system is powered by a 24V 60\,Ah LiFePO\textsubscript{4} (Lithium Iron Phosphate) battery, which feeds into a power distribution board. From this central hub, power is distributed across three primary branches to support all subsystems. One branch supplies power to the RoboClaw motor controller, which drives two MP26 MobilePower motors. These motors provide encoder feedback that enables closed-loop control for precise motion execution.

Another branch routes power through an 24\,V-to-5\,V DC converter. This regulated supply powers both the ESP32 microcontroller and the RadioMaster receiver, ensuring stable operation of control logic and wireless communication. A third branch delivers power directly to a 4-channel relay module. This module is responsible for activating the pneumatic solenoids and features opto-isolated relay circuits that are triggered by the general pin input/ouput (GPIO) outputs from the ESP32. These send signals to activate or deactivate certain modules. 

The ESP32 is a microcontroller developed by Espressif Systems and serves as the central control unit of the system and the main microprocessor on the robot. It communicates bidirectionally with the RoboClaw via UART and interprets encoder signals through its GPIO pins. The Universal Asynchronous Receiver-Transmitter (UART) is a hardware device within the ESP32 used for asynchronous serial communication \cite{microkit-gpio-basics}. It also receives wireless control commands from the RadioMaster receiver through UART1. Based on these inputs, the ESP32 actuates the appropriate relays to control the pneumatic solenoids. Each of the solenoids used, (Solenoids E4D29-000), is connected to a normally open channel on the relay module, enabling high-side switching with isolated control logic \cite{espressif2022esp32}.

\subsection{Software Architecture}
Building off of the electronics system, software was developed to control the electronics system for the AgriCruiser. The software architecture for the AgriCruiser is designed to unify diverse hardware components under a coherent and extensible control framework.  At its core, the ESP32 microcontroller coordinates the driving, actuation, and communication tasks  through a modular software design that prioritizes reliability, maintainability, and scalability. 



\begin{figure}[h!]
    \centering
    \includegraphics[width=1\linewidth]{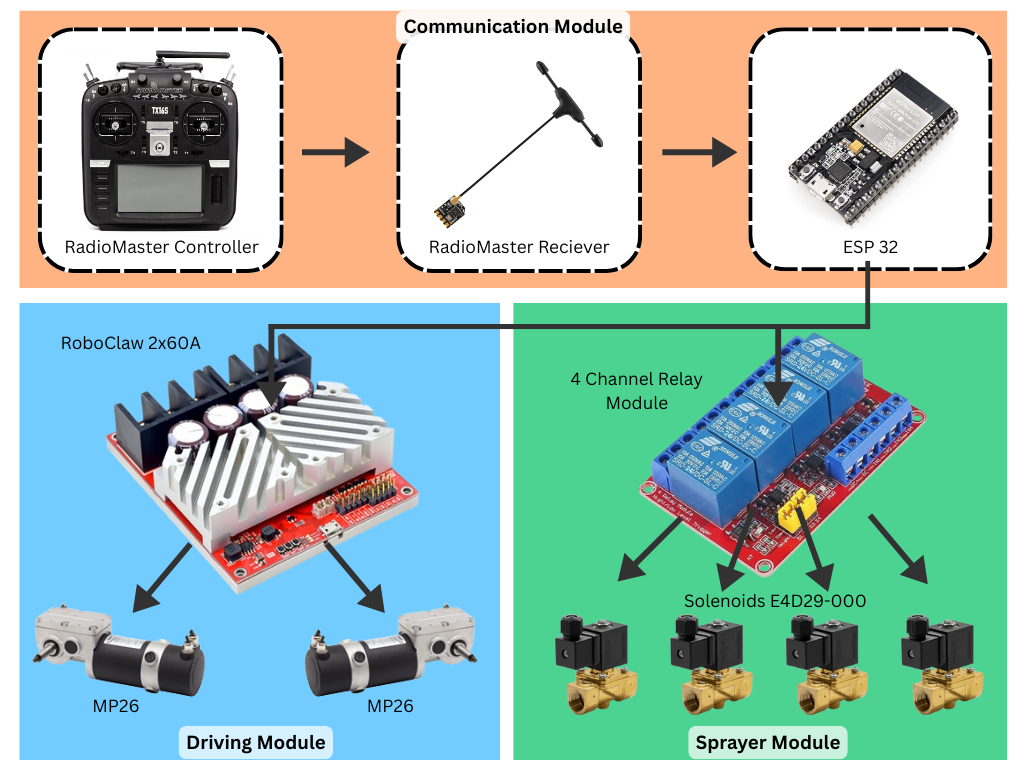}
    \caption{There are three main modules that make up our control system. These are the Communication, Driving, and Sprayer Module. Each focuses on controlling a specific part of the AgriCruiser.}
    \label{fig:system_diagram}
\end{figure}

\subsection{Modular Design \& Communication}
The software architecture employs three primary modules that operate independently while maintaining coordinated communication (see Figure~\ref{fig:system_diagram}). This modular approach provides clear fault isolation, simplified debugging, and seamless integration of future subsystems—critical requirements for agricultural robotics where diverse field tasks demand specialized sensing and actuation capabilities. 

The communication flow begins with the operator’s RadioMaster controller where it sends commands based on joystick and switch inputs. Commands are received by the onboard RadioMaster receiver and passed to the ESP32 via the CRSF. The CRSF is the telemetry protocol that can be used for communication between compatible RC transmitters \cite{px4-crsf-telemetry}. The ESP32 is where inputs are interpreted into driving directions and valve commands for activating the solenoids. The ESP32 processes this information and then issues motor instructions to the RoboClaw via UART and toggles relay channels for solenoid activation. The RoboClaw uses its built in PID (proportional-integral-derivative) feedforward controller to help maintain speeds that we want our robot to perform. \cite{boghossian_brown_zak_pid_control}. These values were found through the built-in PID tuner for the RoboClaw. We will get into the steps on how the ESP32 handles this information later. Telemetry collection operates continuously, gathering encoder counts, voltage levels, and motor speeds from the RoboClaw. This data can be transmitted to remote ESP32 nodes via ESP-NOW wireless protocol, supporting real-time monitoring, debugging, and higher-level coordination during field operations.

Each module operates as an independent unit while maintaining communication with the central controller, ensuring that new functionality can be added in a scalable manner. With this structure in place, the AgriCruiser achieves the intended goals of flexibility and functionality. Subsystems can be maintained individually, new hardware can be added seamlessly, and the same framework can be adapted to support a variety of future agricultural research tasks.

\subsubsection{Control Algorithm Implementation}
Utilizing the control system, the flow of data is understood and can be translated towards writing software to help manage the information. Within the ESP32, the code that the ESP32 utilizes follows the logical diagram below. Through this loop, we are able to handle the communications between the controller at the control center and the robot, controlling the robot accurately. 

\begin{algorithm}[h]
\caption{ESP32 Robotic Control Loop}
\label{alg:control}
\begin{algorithmic}[1]
    \State Initialize UART interfaces, GPIO outputs, and ESP-NOW communication
    \While{system is active}
        \State Read Receiver input frame via CRSF (Communication Module)
        \State Normalize joystick and switch channels for solenoids
        \State Compute left/right motor speeds using differential drive logic
        \State Send speed commands to RoboClaw motor driver (Driving module)
        \State Toggle relay channels for solenoids based on switch inputs (Sprayer Module)
        \State Collect encoder and voltage feedback from RoboClaw
        \State Transmit telemetry to remote ESP32 node via ESP-NOW // this is optional (useful for debugging)
    \EndWhile
\end{algorithmic}
\end{algorithm}

The robotic control loop implemented on the ESP32 follows this structured process that allows the AgriCruiser to perform its tasks. The algorithm (see Algorithm~\ref{alg:control}) begins with an initialization phase, where the ESP32 configures its UART interfaces for serial communication, sets up GPIO pins for controlling the relay, and establishes wireless communication through ESP-NOW \cite{espressif_arduino_esp32_espnow}. This ensures that all critical peripherals—such as the RoboClaw motor driver, CRSF receiver, and relay board are ready.

Once initialized, the ESP32 enters a continuous loop where it processes real-time control inputs. First, the system listens for bytes transmitted via the CRSF protocol. This data contains joystick and switch states, which are decoded and normalized into numerical values. Joystick axes are mapped to throttle and steering commands, while dedicated switch channels are reserved for solenoid actuation. This normalization step applies deadzone filtering to eliminate noise and prevent unintended robot movement.

Next, ESP32 computes the motor speed values using differential drive logic. This translates throttle and steering inputs into left and right wheel commands. These commands are sent to the RoboClaw motor driver via UART, which in turn powers the two motors with speed-based feedback control. At the same time, GPIO pins linked to the relay module are toggled depending on switch inputs, enabling independent control of pneumatic solenoids for spraying.

Finally, telemetry and feedback are essential for maintaining stable performance. The ESP32 collects encoder counts, motor speeds, and battery voltage from the RoboClaw, ensuring that the robot’s motion corresponds accurately to the input commands. This information can be transmitted to a secondary ESP32 node via ESP-NOW, providing a wireless feedback channel that is especially useful for debugging or logging during field trials.

In summary, the software architecture provides a robust and adaptable control system for robotic applications involving motorized movement and pneumatic actuation. By integrating real-time wireless communication, closed-loop motor control, and flexible input handling from both RC transmitters and remote nodes, the system is well suited for deployment in remote or semi-autonomous field robots. The modular design ensures that it can be extended to accommodate additional sensors or actuators as mission requirements evolve.

\section{Experiments \& Results}
\label{sec: experiments}
Following the design and manufacturing process, a variety of experiments were conducted at the University of California, Los Angeles (UCLA), US to validate the performance of the robot before deployment in Fargo, North Dakota, US for actual field trials. The experiment results at UCLA confirmed that the robot could reliably navigate various terrain surfaces, ensuring a stable mechanical structure and a robust electronic system. After validation, the robot was secured in a wooden crate and shipped to Fargo. As a result, terrain experiments were conducted both at the UCLA campus and at the Fargo farm site. All farm site evaluations, including robot traversal and herbicide application performance, were conducted in Fargo by combined research teams from UCLA and North Dakota State University (NDSU). 

\subsection{Terrain}
\label{subsec: terrain}
Field experiments were conducted under various surface conditions to evaluate the performance of the robot and its adaptability. The test terrains included concrete floor, asphalt road, unpaved gravel road, grass, wet soil, and dry soil. Each terrain was chosen to evaluate critical performance factors, including mobility and steering accuracy. The selected condition often represents typical agricultural terrains, thereby providing a realistic evaluation of the robot's reliability.

Before field testing, torque requirements for traversing various terrains were calculated to ensure a high success rate during the experiments. The result of the calculations is summarized in Table~\ref{tab: terrain_torque}

\begin{table}[h!]
    \centering
    \begin{tabular}{|c|c|}
        \hline
        \textbf{Surface Conditions} & \textbf{Required Torque} \\
        \hline Concrete & 12.16 Nm  \\ 
        \hline Rough Paved Road & 12.42 Nm  \\ 
         \hline Gravel & 12.95 Nm  \\ 
        \hline Grass & 14.21 Nm  \\ 
        \hline Dry Hard Soil & 13.81 Nm  \\ 
        \hline Wet Saturated  Soil & 15.55 Nm  \\ 
        \hline
    \end{tabular}
    \caption{Total torque required to overcome terrain}
    \label{tab: terrain_torque}
\end{table}

\begin{figure}[h!]
    \centering
    \includegraphics[width=\linewidth]{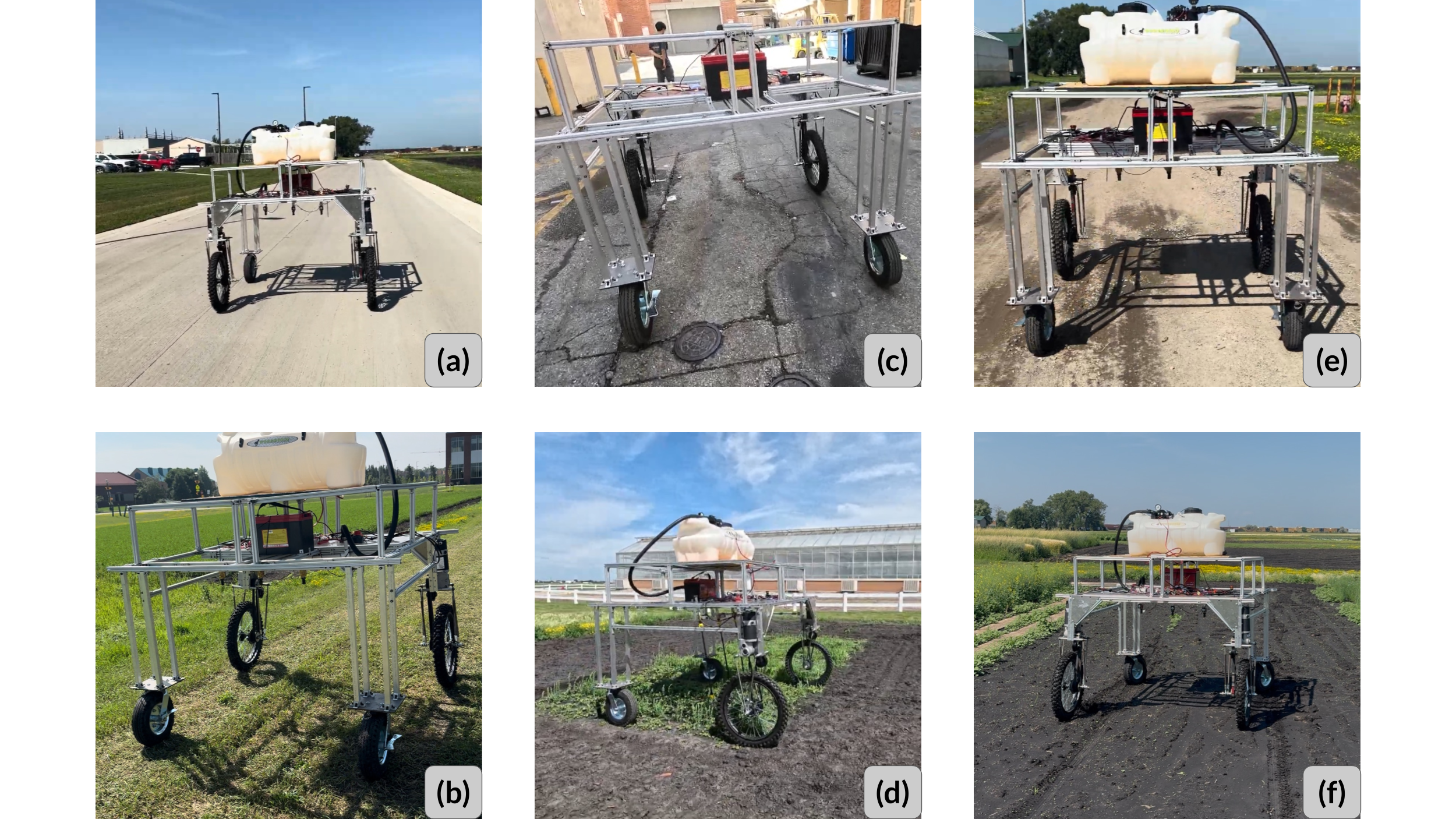}
    \caption{The AgriCruiser on different terrains: (a) concrete, (b) grass, (c) paved road, (d) dry soil, (e) unpaved/gravel road, and (f) wet soil}
    \label{fig: terrain}
\end{figure}

As described in Subsection~\ref{subsec:transmission_system}, the selected motor provides a continuous torque of up to 31.42 Nm for 15 minutes. Since the AgriCruiser employs two motors in its transmission system, the total available continuous torque is 62.82 Nm. The torque output significantly exceeds the minimum torque requirements for agricultural terrains, as mentioned in Table~\ref{tab: terrain_torque}. To validate these calculations, field experiments were conducted in Los Angeles, California, and Fargo, North Dakota. The result of the experiment also confirms that the robot traversed all tested terrains without difficulty, as demonstrated in Figure~\ref{fig: terrain}a - ~\ref{fig: terrain}f

Although no exact quantitative data were collected on the robot's motion across different terrains, the results can be qualitatively observed from Figure~\ref{fig: terrain}a - ~\ref{fig: terrain}f, which shows that the robot moved smoothly without any issues. No corrective intervention was required to complete any run. A major limitation of the previous AgriCruiser design, as well as many other agricultural robots, was that their wheels often get stuck in the soil. This is mainly due to the loose and deformable soil, which increases sinkage and lack of traction. Consequently, when the robot came into contact with the soil, the wheels often stalled, and the robot was unable to move, as was clearly observed in the previous AgriCruiser design. In contrast, the new proposed AgriCruiser design demonstrated robust performance, traversing all the different terrains without slowing down or showing any signs of wheel entrapment. Testing across multiple terrains reflects real agricultural settings, where robots often transition between compact, unpaved paths, grassy areas, and fields with varying soil moisture levels. The caster wheel can be locked or unlocked to control its swiveling motion, preventing unexpected turns of the caster wheel. Locking the rear casters reduced small heading perturbations on uneven ground during long, straight passes, limiting unintended lateral drift within narrow rows. On rougher patches and potholes, the locked setting additionally improved directional stability.

Even without quantitative logs, the absence of stalls or sinkage across the full terrain set provides a practical confirmation that the two-wheel front drive and tire selection meet the tractive demands envisioned in Subsection~\ref{subsec:transmission_system}.

\subsection{Multiple Crops}
\label{subsec: multiple_crops}
A key design objective of the AgriCruiser is its highly flexible and reconfigurable chassis, allowing operation across various crop types and sizes. To demonstrate this flexibility, field experiments were conducted in Fargo, North Dakota. 

\begin{figure}[h!]
    \centering
    \includegraphics[width=\linewidth]{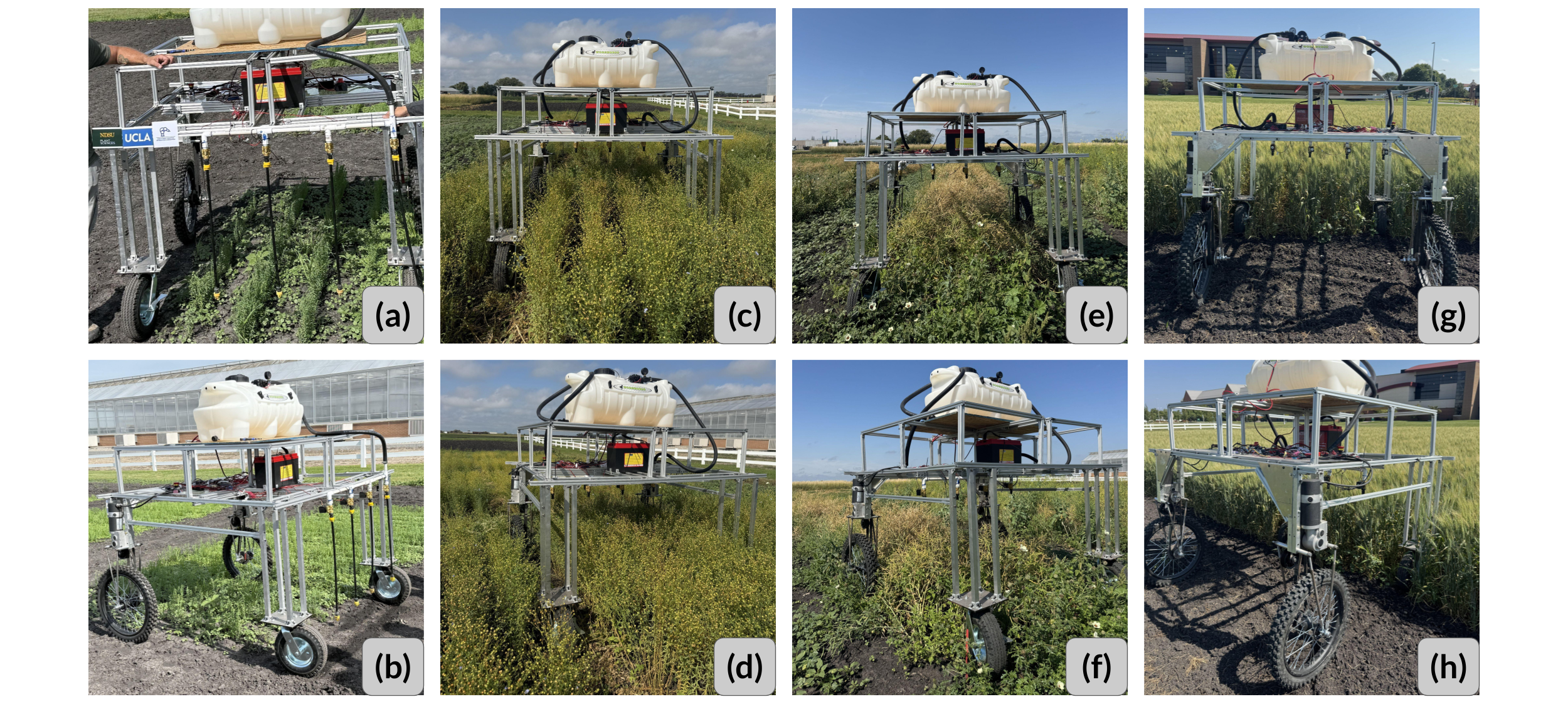}
    \caption{Robot traversing over various crop types with different growth stages (a-b) early flax crop, (c-d) mature flax crop, and (e-f) mature canola crop, and (g-h) mature wheat crop}
    \label{fig: crop}
\end{figure}

In mid-June, the AgriCruiser was tested in a flax field during the early growth stage, when the crop height was low, approximately 0.08-0.13 m. In this trial, the robot successfully traversed over the crop rows and performed weed-spraying tasks without any difficulties, as shown in Figure~\ref{fig: crop}a - ~\ref{fig: crop}b.

Approximately 2.5 months later, at the end of August, the AgriCruiser was tested again in the same flax field during the maturation stage, when the crop height had grown significantly. With its over-the-row design and height configuration, the robot was still able to traverse the crop rows with enough clearances. During this time, the robot was also tested in canola and wheat fields, each with different crop sizes and row layouts. In a canola field, the robot demonstrated smooth traversal, confirming its ability to operate effectively across multiple crop types, as demonstrated in Figure~\ref{fig: crop}a - ~\ref{fig: crop}h. 

As shown in Figure~\ref{fig: crop}g - ~\ref{fig: crop}h, the robot was placed at the entrance of the wheat field, where both the crop height and row width were observed to be within the robot’s clearance. These observations confirm that, if deployed, the robot would be capable of traversing the wheat field without any issues. At maturity, the typical plant heights for flax, canola, and wheat are approximately 0.76 m, 0.89 m, and 0.81 m, respectively. Since all crops remain below the robot clearance of 0.94 m, the platform is capable of traversing these crops even at the harvest stage. Sufficient clearance is an important consideration, as the robot can potentially be used for different agricultural applications such as seeding, nutrient delivery, and harvesting. If the crop height at maturity remains below the clearance threshold, then all earlier growth stages of the crop will also fit under the robot. As a result, the robot can be utilized throughout the entire crop cycle, from the initial planting stage to harvest. 

For these experiments, the robot's track width was reconfigured to accommodate the field-specific row layout. These results confirm that the reconfigurable chassis of the AgriCruiser makes it suitable for deployment across different crops with various dimensions and layouts.  

\subsection{Compact Driving}
\label{compact_drving}

The purpose of the differential drive is to enable the AgriCruiser to perform in-place turning and pivot turning. This ability is essential in agricultural settings, where crop rows are typically planted closely to maximize land use. If the robot required a large turning radius, it would risk damaging surrounding plots. Therefore, it is crucial to design the robot so that it can navigate effectively within compact spaces.

According to the kinematic model presented in Subsection~\ref{subsec:differential_drive_kinematic_model}, the turning radius is defined as the distance from the midpoint between the two driven wheels to the center of rotation. In this pivot-turning experiment, one wheel acts as the center of rotation and remains stationary, while the other wheel rotates around. Thus, the turning radius will be from the midpoint to the stationary wheel, which is half of the track width. 
 
\begin{figure}[h!]
    \centering
    \includegraphics[width=\linewidth]{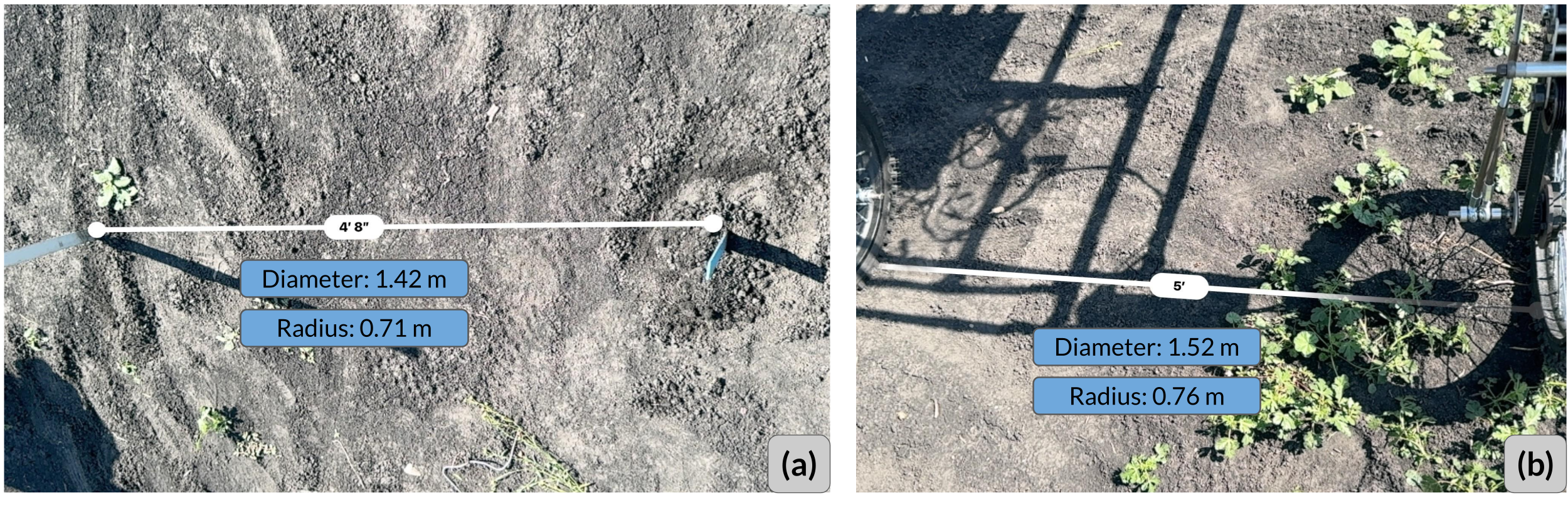} 
    \caption{Measured turning diameter when performing a pivot-turn with (a) 1.42 m track width  and (b) 1.52 m track width}
    \label{fig: turning_diameter}
\end{figure}

Pivot turning experiments were performed with track widths of 1.42 m and 1.52 m. The corresponding turning radii were 0.71 m and 0.76 m, respectively. These results are consistent with the kinematic model, confirming that during pivot turning, the turning radius equals half the track width. Data are shown in Figure~\ref{fig: turning_diameter}a - ~\ref{fig: turning_diameter}b. The robot consistently executed in-place turns without encroaching into adjacent rows.

\subsection{Weed Management Performance}
\label{subsec: weed_managment}
In all the previous experiments, a consistent result was the robot’s ability to traverse different farm sites, overcome uneven terrain, and adapt to various crop types. These outcomes demonstrate the mobility and flexibility of the system. However, in this experiment, the focus shifted to determine whether the robot could efficiently perform its designed agricultural task. While seamless locomotion is essential, the ability to carry out functional field operations is equally important. Since the AgriCruiser was specifically intended for weed management hence the weed control performance was evaluated. 

Although the robot can move smoothly on its own, the integration of additional components, such as weed-spraying nozzles or a mechanical uprooting system, can affect overall performance. For instance, adding a payload to the system may affect the kinematic model and mobility, or additional electrical components may increase hardware issues. Therefore, it is essential to verify that the robot can operate seamlessly when integrated with task-specific tasks.

Weeds such as pigweed and Venice mallow steal nutrients, water, and light from plants, resulting in yield loss if not properly controlled. In this experiment, the AgriCruiser was tested not only for spraying herbicide but also for its effectiveness in spraying. Precise spraying requires stable locomotion from the transmission system, as well as reliable electronic control of the solenoid valves. 

\begin{figure}[h!]
    \centering
    \includegraphics[width=\linewidth]{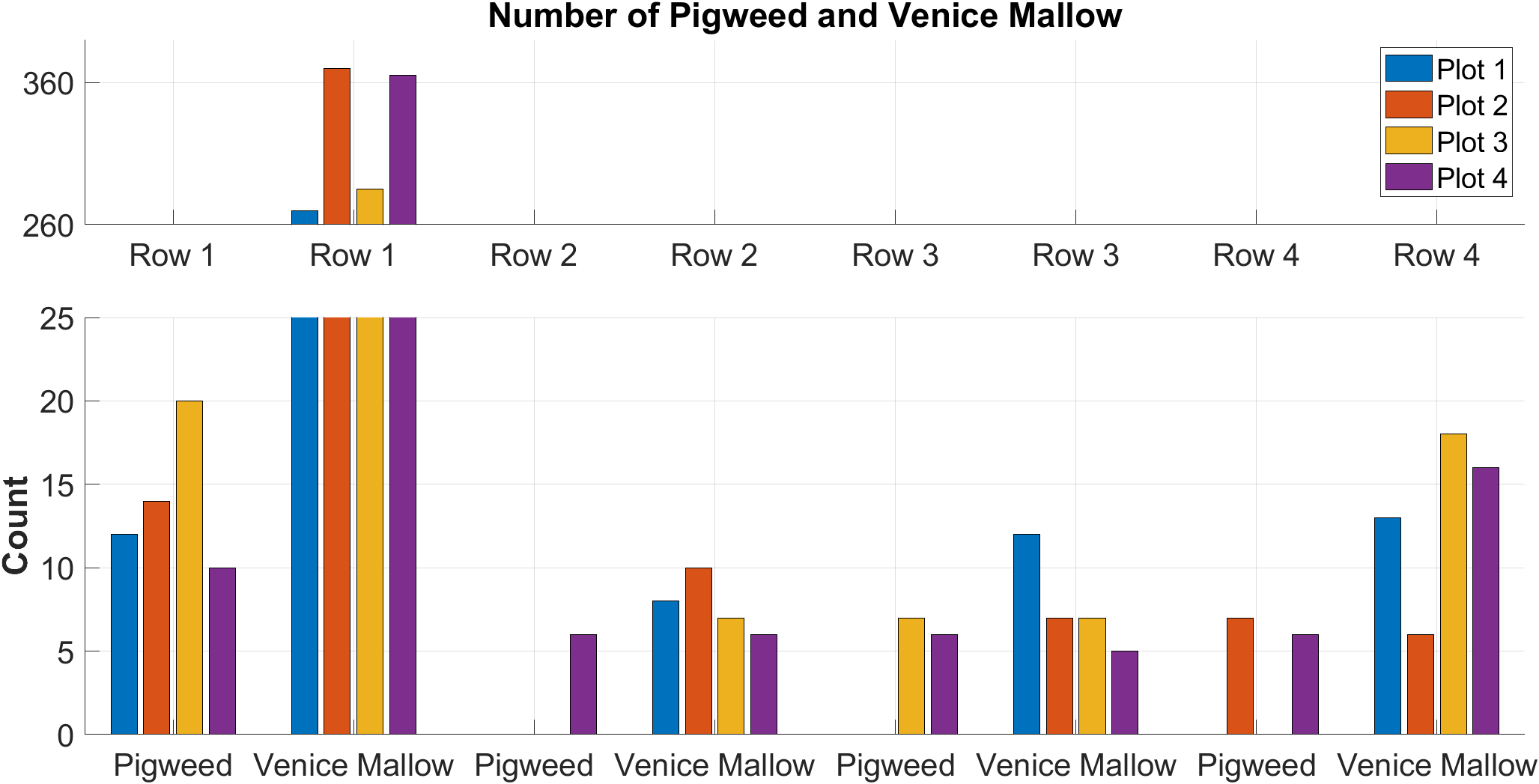}
    \caption{Weed management effectiveness between herbicide control in row 2-4 and manual control in row 1, with row 1 having significantly higher pigweed and Venice mallow weed density}
    \label{fig: Weed_management_graph}
\end{figure}

The field layout consisted of four flax rows, and each row had four plots. The first row (4 plots) was manually controlled using spades, while the remaining three rows (12 plots) were managed by herbicide spraying. In mid-June, the AgriCruiser performed the first round of spraying. The experiment used a mixture of three herbicides: MCPA, Select Max, and Basagran, where 30 milliliters of each herbicide were diluted in 37.85 L of water. However, for the experiment, only 18.93-22.71 L of the solution was added to the tank as the required spray volume was lower than that. A second round was scheduled a month later, but it was unnecessary due to the high effectiveness of the first round. At the end of August, the weed counts were conducted across the 16 plots, and the results are shown in Table~\ref{tab: weed_management_table} and Figure~\ref{fig: Weed_management_graph}

\begin{table}[h!]
    \centering
    \begin{tabular}{|p{1cm}|p{1.2cm}|p{1.2cm}|p{1.2cm}|p{1.2cm}|p{1.2cm}|p{1.2cm}|p{1.2cm}|p{1.2cm}|}
        \hline
        & \multicolumn{2}{c|}{\textbf{Row 1}}  & \multicolumn{2}{c|}{\textbf{Row 2}}  & \multicolumn{2}{c|}{\textbf{Row 3}}  & \multicolumn{2}{c|}{\textbf{Row 4}}   \\
        \hline 
        & \multicolumn{2}{c|}{Manually cleaned} & \multicolumn{6}{c|}{Sprayed} \\ 
        \hline & Pigweed & Venice Mallow & Pigweed & Venice Mallow  & Pigweed & Venice Mallow & Pigweed & Venice Mallow \\
        \hline Plot 1 & 12 & 270 & 0 & 8 & 0 & 12 & 0 & 13 \\
        \hline Plot 2 & 14 & 370 & 0 & 10 & 0 & 7 & 0 & 6 \\
        \hline Plot 3 & 20 & 285 & 0 & 7 & 0 & 7 & 0 & 18\\
        \hline Plot 4 & 10 & 365 & 0 & 6 & 0 & 5 & 0 & 16\\
        \hline
    \end{tabular}
    \caption{Number of pigweed and Venice mallow two months post-herbicide spraying}
    \label{tab: weed_management_table}
\end{table}

\begin{figure}[h!]
    \centering
    \includegraphics[width=\linewidth]{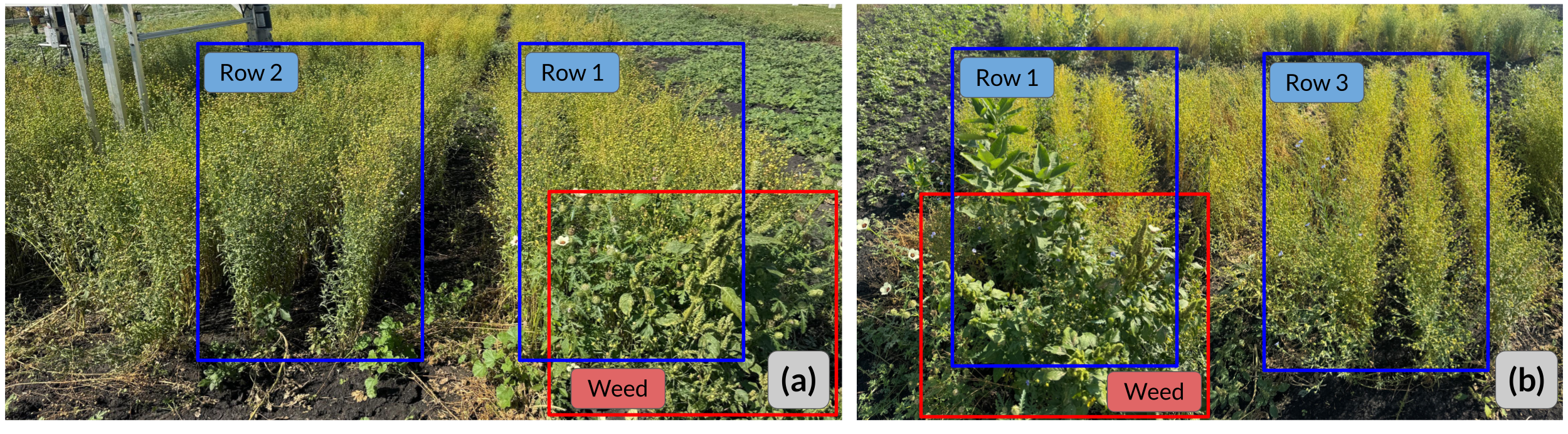}
    \caption{Comparison of weed density between (a) Front side of Flax Row 1 and Flax Row 2, with higher weed density in Row 1 highlighted in Red, and (b) Back side of Flax Row 1 and Flax Row 3, with higher weed density in Row 1 highlighted in Red}
    \label{fig: Weed_management_performance}
\end{figure}

The data show that the number of pigweed and Venice mallow in the manually controlled plots (row one) was significantly higher compared to the rows sprayed with herbicide. Specifically, row one contained more than 42 times more weeds than rows two and three, and about 24 times more than row four. Figure~\ref{fig: Weed_management_performance} shows a clear contrast in the presence of weeds between row one and the other rows, with row one clearly having a higher weed density than the other two rows. 

These results demonstrate that a single round of robotic spraying was much more effective than manual weeding, confirming that the AgriCruiser can achieve precise and efficient weed management. 

\subsection{Crop and Soil Safety}
\label{subsec: crop_and_soil_safety}
Crop safety is a crucial factor for the implementation of autonomous and large robots in agricultural settings, as farmers are afraid to risk yield loss due to crop damage. As mentioned above, crop rows are often planted in close proximity to maximize land use. Therefore, it is essential to evaluate the robot’s capability to traverse between rows with minimal crop damage. 

In this experiment, row one was manually cleaned using garden hoes, while rows two through four were sprayed with herbicide. Crop safety during weed removal is difficult to guarantee under manual methods, particularly in dense or intra‐row conditions. Empirical studies \cite{zheng2025design} show that when plant spacing is small or when operators are working quickly, spatial errors of only a few centimeters can produce a non-trivial crop injury rate (stem or leaf damage). Moreover, even where physical contact is avoided, weeds close to crops compete for light, water, and nutrients; thus mis-timed or imprecise removal amplifies yield loss. Table~\ref{tab: damaged_plants_table} summarizes damaged crops after multiple traversals of the AgriCruiser through the field, compared to manual control. 


\begin{table}[h!] 
\centering 
\begin{tabular}{|c|c|c|c|c|} 
\hline 
& {\textbf{Row 1}} & \textbf{Row 2} & \textbf{Row 3} & \textbf{Row 4} \\ 
\hline 
& Manually cleaned & \multicolumn{3}{c|}{Sprayed} \\ 
\hline 
Plot 1 & 100 & 20 & 15 & 45 \\ 
\hline 
Plot 2 & 85 & 5 & 70 & 50 \\ 
\hline 
Plot 3 & 38 & 5 & 13 & 55 \\ 
\hline 
Plot 4 & 100 & 25 & 25 & 43 \\ 
\hline 
Total & 323 & 55 & 123 & 195 \\ 
\hline 
\end{tabular} 
\caption{Number of plants damaged during the robot traversal} 
\label{tab: damaged_plants_table} 
\end{table}

The data show that the rows traversed by the AgriCruiser experienced less crop damage compared to the rows manually controlled with garden hoes. Row one has 5.8 times more damaged plants than row two, 2.5 times more than row three, and about 1.65 times more than row four. The over-the-row design, flexible chassis, narrow transmission and sufficient clearance enable the robot to maintain a safe distance from the crop, thus reducing the rate of crop damage. These results confirm both the effectiveness and the safety of deploying the AgriCruiser in field operations.

\section{Conclusion and Future Works}
\label{sec: conclusion}
As robotics and autonomous vehicles continue to expand across various industries, smart farming has become an important development within the agriculture industry. This paper addresses several challenges in the agricultural sector and highlights the importance of implementing smart farming solutions. The AgriCruiser is the first stage of a long-term project with the aim of tackling these challenges. With its dynamic and flexible platform design, the robot demonstrated reliable operation across different farm fields with various crop types and sizes. As an open-source platform, the AgriCruiser is designed for broad adoption, allowing farmers worldwide to implement the robot without being limited to specific crops. Field experiments further validated the integration of a herbicide spraying subsystem, showing efficient weed management. This success establishes a foundation for extending the platform to incorporate other agricultural subsystems, such as seeding, nutrient delivery, and harvesting, thereby advancing the concept of a multipurpose and sustainable robot in the agricultural industry.

With its dynamic and flexible platform design, the robot demonstrated reliable operation across different farm fields with various crop types and sizes. As shown in Subsection~\ref{subsec: terrain}, the experimental results confirm that the robot can navigate reliably across various terrain conditions, including challenging surfaces such as wet soil. Unlike many small-scale vehicles, which often get stuck in those conditions, the AgriCruiser successfully traversed wet soil two days after heavy rain in Fargo, thus validating its ability to operate effectively in adverse field environments. As an open-source platform, the AgriCruiser is designed for broad adoption, allowing farmers worldwide to implement the robot without being limited to specific crops. The robot successfully traversed the flax, canola, and wheat fields, as presented in Subsection~\ref{subsec: multiple_crops}. The results indicate that the platform was able to move through these fields while minimizing crop damage, highlighting its reliability to operate at different farm sites, including more compact field layouts. Weed control experiments were performed manually by hand controlling one row while applying herbicide to three rows for comparison. As shown in Subsection~\ref{subsec: weed_managment}, the rows sprayed with herbicide contained significantly lower weed density than the manually controlled row. These results validate the effectiveness of the integrated spraying module subsystem, demonstrating its capability to incorporate additional functional modules. This success establishes a foundation for extending the platform to perform additional farming operations, such as seeding, nutrient delivery, and harvesting, thus advancing the concept of a sustainable multipurpose robot in the agricultural industry.

Based on the experimental results, the AgriCruiser has been demonstrated to be a reliable platform to operate
effectively in agricultural environments. However, although the AgriCruiser is capable of traversing extremely rough terrain, the robot's motion is not smooth due to the lack of a suspension system. In the next iteration, a suspension system would be developed to absorb shocks when operating on uneven ground, thus improving motion over rough terrain smoothly. The next stage of development also focuses on autonomous navigation through the integration of AgroNav, an autonomous navigation framework for agricultural vehicles that utilizes semantic segmentation and semantic line detection \cite{agronav}. The implementation of AgroNav will allow the AgriCruiser to navigate autonomously on fields, further enhancing its adaptability and reducing the need for human supervision.

\subsection*{Code and Data Availability}
All design files, source code, and documentation for the AgriCruiser are available at \url{https://github.com/StructuresComp/agri-cruiser/}.

\subsubsection*{Acknowledgments}
The authors gratefully acknowledge support from USDA grants \#2024-67021-42528, \#2022-67022-37021, and \#2021-67022-34200.

\end{document}